\documentclass{article}
\usepackage{spconf,amsmath,graphicx}

\usepackage{multirow}     
\usepackage{booktabs}     
\usepackage{arydshln}     
\usepackage{makecell}     
\usepackage{array}        
\usepackage{amssymb}      
\usepackage{bbding}
\usepackage{pifont}
\usepackage{utfsym}       

\usepackage{pgfplots}     
\usepackage{subfigure}    
\usepackage{tikz}         
\pgfplotsset{compat=1.18} 
\usepgfplotslibrary{groupplots} 
\usepgfplotslibrary[groupplots]
\usetikzlibrary{pgfplots.groupplots}
\usetikzlibrary[pgfplots.groupplots]
\usepackage{hyperref}

\hypersetup{hidelinks,
	colorlinks=true,
	allcolors=black,
	pdfstartview=Fit,
	breaklinks=true}

\title{
Combining Recent Advances of CTC \\for Speech Translation and Speech Recognition
}
\title{Bridging the Gaps of Both Modality and Language: Synchronous Bilingual CTC for Speech Translation and Speech Recognition}
\name{\begin{tabular}{c}
    Chen Xu\textsuperscript{1}, Xiaoqian Liu\textsuperscript{2}, Erfeng He\textsuperscript{2}, Yuhao Zhang\textsuperscript{2} \\
    Qianqian Dong\textsuperscript{3}, Tong Xiao\textsuperscript{2,4}, Jingbo Zhu\textsuperscript{2,4}, Dapeng Man\textsuperscript{1}, Wu Yang\textsuperscript{1*}\footnotemark[1]
\end{tabular}}
\address{\textsuperscript{1}Harbin Engineering University\\
\textsuperscript{2}School of Computer Science and Engineering, Northeastern University, Shenyang, China\\
\textsuperscript{3}ByteDance \\
\textsuperscript{4}NiuTrans Research, Shenyang, China }
\begin{document}
\ninept
\maketitle
\renewcommand{\thefootnote}{\fnsymbol{footnote}}
\footnotetext[1]{Corresponding author: yangwu@hrbeu.edu.cn}
\renewcommand{\thefootnote}{\arabic{footnote}}
\begin{abstract}


In this study, we present synchronous bilingual Connectionist Temporal Classification (CTC), an innovative framework that leverages dual CTC to bridge the gaps of both modality and language in the speech translation (ST) task.
Utilizing transcript and translation as concurrent objectives for CTC, our model bridges the gap between audio and text as well as between source and target languages.
Building upon the recent advances in CTC application, we develop an enhanced variant, BiL-CTC+, that establishes new state-of-the-art performances on the MuST-C ST benchmarks under resource-constrained scenarios. 
Intriguingly, our method also yields significant improvements in speech recognition performance, revealing the effect of cross-lingual learning on transcription and demonstrating its broad applicability.
The source code is available at \href{https://github.com/xuchennlp/S2T}{https://github.com/xuchennlp/S2T}.

\end{abstract}
\section{Introduction}
\label{sec:intro}

Automatic speech recognition (ASR) and speech translation (ST) have undergone remarkable evolution in recent years. 
Current end-to-end systems consistently outperform their traditional hybrid and cascaded counterparts in both efficiency and performance \cite{Gulati_ISCA2020, xu2023recent}. 
Despite these progresses, achieving stable convergence still remains a challenge due to the inherent complexities arising from the modality gap between audio and text.
To mitigate this, Connectionist Temporal Classification (CTC) \cite{Graves_ACL2006} has emerged as a widely-used auxiliary training objective \cite{watanabe2017hybrid}, which has demonstrated a notable improvement in stabilizing model convergence and enhancing performance, especially in the more challenging ST task \cite{bahar2019comparative}.

CTC is typically implemented on top of the encoder, where it predicts the corresponding transcript of the input audio. 
This mechanism offers a more lightweight and direct solution compared to alternative techniques like pre-training \cite{bansal2018pre} or data augmentation \cite{pino_corr2019}. 
Nevertheless, this approach has its limitations.
In particular, its narrow focus on the transcription can potentially interfere with the semantic understanding that is critical in the ST task \cite{Xu_ACL2021}.

Researchers propose a stacked design for ST that decouples the encoding into an acoustic encoder and a textual encoder, where only the former is supervised by the CTC loss\cite{Xu_ACL2021,nast}. 
Considering the gaps of both modality and language in the ST task, the following study further enhances it by introducing another cross-lingual CTC (XCTC), which guides the textual encoder to predict target translation \cite{nast}, as illustrated in Figure \ref{fig:progressive}.
This approach offers comprehensive guidance from the bilingual text in a progressive manner.
We refer to it as bilingual CTC (BiL-CTC).

\definecolor{qblue}{RGB}{174,208,238}
\definecolor{qpurple}{RGB}{187,161,203}
\definecolor{qgreen}{RGB}{177,213,200}
\definecolor{qorange}{RGB}{243,166,148}
\definecolor{qpink}{RGB}{249,199,207}
\definecolor{qyellow}{RGB}{254,208,129}

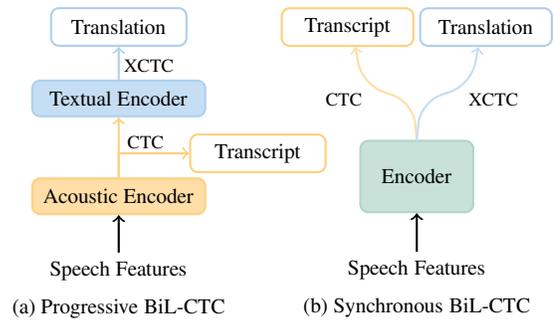
\begin{figure}
    \centering
\subfigure{

    \rotatebox{10}{}
    \begin{minipage}[t]{0.4\columnwidth}
    \label{fig:progressive}
        \centering
        \begin{tikzpicture}
            \tikzstyle{textonly} = [font=\footnotesize, align=center];
            \tikzstyle{encoder} = [rectangle,draw,minimum width=1.5cm,rounded corners=3pt,align=center,inner sep=3pt,minimum height=0.5cm,font=\footnotesize];
            \tikzstyle{sequence} = [rectangle,rounded corners=3pt,minimum width=1.8cm,align=center,inner sep=1pt,minimum height=0.5cm,font=\footnotesize];
            \tikzstyle{arrow} = [font=\Large,align=center];

            \node[textonly] (inputb) at (0,0) {Speech Features};

            \node[textonly, font=\footnotesize] (title) at (0,-0.5) {(a) Progressive BiL-CTC};
            
            \node[encoder, draw=qyellow, thick, fill=qyellow!70, minimum height=1.5em, minimum width=2.3cm] (aencoder) at ([shift={(0,0.75)}]inputb.north) {Acoustic Encoder}; 

            \node[sequence, draw=qyellow, thick] (source) at ([shift={(1.85,0.32)}]aencoder.north) {Transcript}; 
            
            \node[encoder, draw=qblue, thick, fill=qblue!70, minimum height=1.5em, minimum width=2.3cm] (tencoder) at ([shift={(0,1.05)}]aencoder.north) {Textual Encoder}; 

            \node[sequence, draw=qblue, thick,minimum width=1.8cm] (target) at ([shift={(0,0.7)}]tencoder.north) {Translation}; 
            

            \node[textonly, font=\scriptsize] (ctc) at ([shift={(-0.58,0.15)}]source.west) {CTC}; 

            \node[textonly, font=\scriptsize] (xctc) at ([shift={(0.4,-0.25)}]target.south) {XCTC}; 

            \draw[->,thick, draw=qyellow] (aencoder.north) -- (tencoder.south);
            \draw[->,thick, draw=qyellow] (aencoder.north) -- ([shift={(0,0.32)}]aencoder.north) -- (source.west);
            \draw[->,thick, draw=qblue] (tencoder.north) -- (target.south);
            \draw[->, thick] (inputb.north) -- (aencoder.south);
            
            
        \end{tikzpicture}
    \end{minipage}
}
\subfigure{
    \begin{minipage}[t]{0.4\columnwidth}
    \label{fig:syn}
        \centering
        \begin{tikzpicture}
            \tikzstyle{textonly} = [font=\footnotesize, align=center];
            \tikzstyle{encoder} = [rectangle,draw,minimum width=1.5cm,rounded corners=3pt,align=center,inner sep=3pt,minimum height=0.5cm,font=\footnotesize];
            \tikzstyle{sequence} = [rectangle,rounded corners=3pt,minimum width=1.75cm,align=center,inner sep=1pt,minimum height=0.5cm,font=\footnotesize];
            \tikzstyle{arrow} = [font=\Large,align=center];

            \node[textonly] (input) at (0,0) {Speech Features};
            \node[textonly, font=\footnotesize] (title) at (0,-0.5) {(b) Synchronous BiL-CTC};
            \node[encoder, draw=qgreen, thick, fill=qgreen!70,minimum height=3em] (pencoder) at ([shift={(0,1)}]input.north) {Encoder}; 
            
            \node[sequence, draw=qyellow, thick] (source) at ([shift={(-0.92,1.5)}]pencoder.north) {Transcript}; 
            
            \node[sequence, draw=qblue, thick] (target) at ([shift={(0.92,1.5)}]pencoder.north) {Translation};
            

            \node[textonly, font=\scriptsize] (ctc) at ([shift={(-0.05,-0.7)}]source.south) {CTC};

            \node[textonly, font=\scriptsize] (xctc) at ([shift={(0.1,-0.7)}]target.south) {XCTC};

            \draw[->, thick, rounded corners=8pt, draw=qyellow!70] (pencoder.north) -- ([shift={(-0.05,0.5)}]pencoder.north) -- ([shift={(0.2,-0.5)}]source.south) -- ([shift={(0.1,0)}]source.south);
            
            \draw[->, thick, rounded corners=8pt, draw=qblue!70] (pencoder.north) -- ([shift={(0.05,0.5)}]pencoder.north) -- ([shift={(-0.2,-0.5)}]target.south) -- ([shift={(-0.1,0)}]target.south);

            \draw[->, thick] (input.north) -- (pencoder.south);
            
        \end{tikzpicture}
    \end{minipage}
}
\caption{Two strategies of BiL-CTC.
}
\label{overview}
\end{figure}

In this paper, we present a novel implementation of BiL-CTC that synchronously predicts both transcript and translation, as illustrated in Figure \ref{fig:syn}. 
While the progressive strategy of incrementally learning the cross-modal and cross-lingual prediction may seem intuitively sound, our synchronous approach promotes mutual assistance between bilingual prediction.
This not only facilitates the learning of language-agnostic representations but also enhances the accuracy of semantic understanding.
This strategy is inspired by the paradigm shift from cascaded systems to end-to-end models, where direct semantic comprehension replaces the two-stage pipeline of cross-modal transcription followed by cross-lingual translation.

We further come up with an enhanced variant by combining several recent advancements in CTC application, including InterCTC \cite{Lee_ICASSP2021}, prediction-aware encoding \cite{nast}, and a curriculum-based training strategy \cite{nast}, referred to as BiL-CTC+. 
The resulting speech translation model not only establishes a new state-of-the-art performance on the MuST-C corpora in resource-constrained scenarios but also outperforms most of the methods that rely on abundant external data. 
Intriguingly, we observe that our method consistently improves ASR model performance across both low-resource and high-resource settings. 
To the best of our knowledge, this is the first work to demonstrate that direct cross-lingual modeling can positively impact ASR.

\section{Method}

In this section, we first present bilingual CTC (BiL-CTC), a framework to predict both transcript and translation by dual CTC.
Then we propose two distinct implementation strategies.
Finally, we augment BiL-CTC with recent advances in CTC, resulting in a more powerful variant BiL-CTC+.

\subsection{Bilingual CTC}

Speech translation involves directly translating audio in a source language to text in a target language \cite{berard2016listen}. 
Consider an ST training sample denoted as $(s; x; y)$, where $s$ represents the feature sequence of the audio input, $x$ is its transcription, and $y$ is the corresponding translation in the target language.
The vanilla training objective using cross-entropy loss for an ST model parameterized by $\theta$ based on an encoder-decoder architecture can be formalized as:
\begin{eqnarray}
\mathcal{L}_{\rm CE} = -\log \textrm{P}_{\rm CE} (y | s; \theta)
\end{eqnarray}

However, achieving stable convergence in end-to-end ST remains challenging due to the inherent complexities of cross-modal and cross-lingual mappings \cite{xu2023recent}. 
A well-established solution is to incorporate the CTC loss at the encoder's output to guide cross-modal learning, which has achieved both stable convergence and remarkable improvements \cite{bahar2019comparative}.
This CTC loss can be formalized as:
\begin{eqnarray}
    \mathcal{L}_{\rm CTC} = -{\rm log} \textrm{P}_{\rm CTC}(x|s)
\end{eqnarray}

Recent literature has further evolved this framework to address the specific requirement of ST modeling.
An additional cross-lingual CTC (XCTC) has been introduced with the objective of predicting translations directly within the encoder, thereby enhancing cross-lingual learning capability \cite{nast}.
This approach is orthogonal to the original CTC and contributes to additional performance gains.
The XCTC loss is similarly defined as:
\begin{eqnarray}
    \mathcal{L}_{\rm {XCTC}} = -{\rm log} \textrm{P}_{\rm XCTC}(y|s)
\end{eqnarray}

By integrating CTC and XCTC, the encoder benefits from explicit and direct supervision brought by bilingual prediction.
This serves to mitigate the inherent complexities of modeling burden.
We refer to this framework as bilingual CTC (BiL-CTC).
These auxiliary CTC losses are integrated using respective weights:
\begin{eqnarray}
    \mathcal{L} = \mathcal{L}_{\rm CE} + \alpha \cdot \mathcal{L}_{\rm CTC} + \beta \cdot \mathcal{L}_{\rm XCTC}
    \label{two_loss}
\end{eqnarray}
where $\alpha$ and $\beta$ are the coefficients of the CTC and XCTC losses.

\subsection{Implementation Strategy}

In conventional implementations, CTC loss is typically computed from the output of the encoder's final layer. 
However, within the context of our BiL-CTC framework, this is an open problem. 
We investigate two distinct strategies, as illustrated in Figure \ref{overview}.

\subsubsection{Progressive Method}

In the field of ST, existing studies typically decouple the encoding process into two specialized components in a stacked architecture: an acoustic encoder and a textual encoder \cite{Xu_ACL2021, nast}.
They individually handle cross-modal and cross-lingual modeling. 
BiL-CTC enhances this architecture by assigning the CTC and XCTC losses to the acoustic and textual encoders:
\begin{eqnarray}
    \mathcal{L}_{\rm CTC} &=& -{\rm log} \textrm{P}_{\rm CTC}(x|s; \theta_{ae})  \\
    \mathcal{L}_{\rm XCTC} &=& -{\rm log} \textrm{P}_{\rm XCTC}(y|s; \theta_{ae}, \theta_{te})
\end{eqnarray}
where $\theta_{ae}$ and $\theta_{te}$ denote the parameters of the acoustic and textual encoders, respectively.

Under this strategy, the encoder first learns the mapping from input audio features to a semantic representation in the source language. 
It then understands the semantics of the whole sequence and generates the corresponding representation in the target language.
This progressive strategy mimics the recipe of cascaded speech translation and shows promise for achieving stable and improved convergence.

\subsubsection{Synchronous Method}

While the progressive method offers an intuitively reasonable design, we propose an alternative: synchronous prediction of bilingual CTC. 
Unlike the progressive approach, which explicitly decouples acoustic and textual encoding, the synchronous method concurrently calculates both CTC and XCTC losses with the same representation:
\begin{eqnarray}
    \mathcal{L}_{\rm CTC} &=& -{\rm log} \textrm{P}_{\rm CTC}(x|s; \theta_{e})  \\
    \mathcal{L}_{\rm XCTC} &=& -{\rm log} \textrm{P}_{\rm XCTC}(y|s; \theta_{e})
\end{eqnarray}
where $\theta_{e}$ is the parameter of the encoder.

In this design, the encoder is guided to learn a language-agnostic semantic representation, which is well-aligned with the motivation of end-to-end ST.
Compared to the progressive method, the synchronous strategy stands out for its simplicity and potential mutual enhancement between transcript and translation generation.

\subsection{Recent Advances}

The various application of CTC has recently gained considerable attention in the fields of ASR and ST. 
In the following, we incorporate various recent lightweight and orthogonal innovations into the BiL-CTC framework, which we refer to as BiL-CTC+.

\noindent\textbf{InterCTC}\;\cite{Lee_ICASSP2021}
Given that CTC loss is primarily calculated by the output of the top encoder layer, there exists an inherent risk of insufficient regularization in the lower layers.
To mitigate this, InterCTC introduces auxiliary CTC losses, calculated based on the representations generated by various intermediate layers, as formulated below:
\begin{eqnarray}
    \mathcal{L_{\rm InterCTC}} = -{\rm log} \textrm{P}_{\rm InterCTC}(x|h^l)
\end{eqnarray}
where $h^l$ denotes the output of the encoder layer $l$. 
This ensures that the lower layers receive more direct supervision of CTC, thereby enhancing the regularization effectively.

\noindent\textbf{PAE}\;\cite{nast} 
CTC ignores the interdependence within the whole sequence due to the inherent conditional independence assumption. 
The prediction-aware encoding (PAE) method explicitly incorporates prediction information at the intermediate layer $l$, thus enriching the subsequent encoding process.
This is achieved by weighting the embedding matrix $W$ over the current CTC distribution \cite{nozaki2021relaxing}:
\begin{eqnarray}
    \label{pae}
    \textrm{PAE}(h^l) = h^l + \textrm{P}_{\rm InterCTC}(\pi|h^l) \cdot W
\end{eqnarray}
where the weights $W$ are shared across the whole network.

\noindent\textbf{CLM}\;\cite{nast} 
Direct learning of target translation through CTC is non-trivial.
To alleviate this issue, the curriculum learning mixing (CLM) method offers an adaptive training scheme that complements PAE.
Specifically, CLM leverages ground truth to randomly replace a portion of the incorrect predictions in Eq. \ref{pae}.
This strategy mitigates the side effects of error propagation potentially arising from poor XCTC performance in PAE.
The ground truth of CTC is derived from the best alignment $\pi$ computed by the model:
\begin{eqnarray}
    \hat{\pi} = \arg \max_{\pi \in \Phi(y)} \textrm{P}(\pi | s; \theta^{'})
\end{eqnarray}
where $\theta^{'}$ is the current model parameter. 

\subsection{Inference}

In the field of ASR and ST, inference based on the attention mechanism (Attn-only)  serves as the default method, offering high flexibility and effective contextual modeling. 
Nevertheless, it is prone to error, particularly when managing long sequence prediction. 
Conversely, inference based on CTC (CTC-only) preserves the alignment consistency between the audio and the text.
The re-scoring approach \cite{hybridctc2017} employs a weight parameter, $\lambda$, to combine the individual merits of both CTC and attention-based paradigms via one-pass decoding.
This fusion ensures both alignment consistency and generation accuracy.

\section{Experiments}

\subsection{Datasets and Pre-processing}

We primarily focus on the evaluation of BiL-CTC within the ST task, and then extend it to the ASR task.
We construct the experiments on the MuST-C and LibriSpeech benchmarks, respectively.

\noindent{\textbf{LibriSpeech}} is a publicly available English ASR corpus, containing 960-hour training data \cite{Panayotov_ICASSP2015}. 
To evaluate the performance in both low-resource and high-resource scenarios, we conduct experiments on two distinct sets: a 100-hour clean subset referred to as LibriSpeech 100h, and the entire 960-hour training set referred to as LibriSpeech 960h. 
The development and test data are categorized into clean and other subsets according to the speech quality. 
We report results on all four subsets: dev-clean, dev-other, test-clean, and test-other.
Considering that the ASR dataset does not include the labeled translation, we use the online translator to obtain the German translation as the training objective of the XCTC.

\noindent{\textbf{MuST-C}} is a multilingual ST corpus which extracted from the TED talks \cite{Gangi_NAACL2019}. 
We test our method on all MuST-C v1 corpora, including English (En) to German (De), Spanish (Es), French (Fr), Italian (It), Dutch (Nl), Portuguese (Pt), Romanian (Ro), Russian (Ru).
In addition, we also investigate the results of the distant language pair English-Japanese (En-Ja) corpus in the MuST-C v2 dataset.
We select (and tune) the model on the dev set and report the results on the tst-COMMON set for each language pair.

For pre-processing, we follow the standard recipes in fairseq toolkit\footnote{\href{https://github.com/pytorch/fairseq}{https://github.com/pytorch/fairseq}}. These procedures involve filtering out utterances with more than 3,000 frames or fewer than 5 frames. 
The extraction of 80-channel Mel filter bank features is performed using a 25ms window and a stride of 10ms.
Regarding segmentation, we employ SentencePiece\footnote{\href{https://github.com/google/sentencepiece}{https://github.com/google/sentencepiece}} segmentation with a vocabulary size of 10,000 for LibriSpeech and MUST-C datasets. 
We use a shared vocabulary for the source and target languages for MuST-C v1 corpora, the independent vocabulary for the En-Ja corpus.

\subsection{Model Settings}

All methods are implemented using the fairseq
toolkit.
We use the Adam optimizer and follow the default learning schedule in fairseq. 
We apply dropout with a rate of 0.15 and label smoothing $\epsilon_ {ls} = 0.1 $for regularization.
SpecAugment \cite{Park_ISCA2019} is applied to input speech features for better generalization and robustness.

In the progressive BiL-CTC implementation, the ST model consists of an acoustic encoder with 12 Conformer layers, a textual encoder with 6 Conformer layers, and a decoder with 6 Transformer layers. 
Each layer comprises 512 hidden units, 8 attention heads, and 2048 feedforward sizes.
Conversely, the synchronous method employs a unified encoder consisting of 18 layers by merging the acoustic and textual encoders.
In this way, we maintain the total model parameters at about 150M, ensuring fair comparisons with existing literature.
The ASR model uses the Conformer layer that comprises 256 hidden units and 4 attention heads, and other settings are consistent with those of the ST model.

We employ InterCTC and PAE for both CTC and XCTC.
The weights for CTC and XCTC losses are set to 0.2 and 0.1, and the weights of InterCTC losses are set to half of them.
We apply InterCTC starting from the 6th layer, every three layers.
The mixing ratio for the CLM method is set to 0.1, and the excessive ratio hurts the learning of the decoder.
Note that the CLM method is only used for XCTC learning.

During inference, we average the model parameters on the best 10 checkpoints based on the performance of the development set. 
We use greedy search for CTC-only inference and beam search with a beam size of 5 for Attn-only and re-scoring inference.
The weight of CTC for re-scoring is set to 0.1.
We report case-sensitive SacreBLEU on the MuST-C datasets and word error rate (WER) on the LibriSpeech datasets for standardization comparison across papers.

\subsection{Ablation Study}
\begin{table}[t!]
  \centering
  \footnotesize
  \begin{tabular}{llccc}
    \toprule
    \multirow{2}*{Strategy} & \multirow{2}*{Method} & \multicolumn{3}{c}{Inference}\\
    \cmidrule(lr){3-5} 
    & & CTC & Attn & Re-scoring \\
  
    \specialrule{0em}{1pt}{1pt}
    \midrule
    \multirow{6}{*}{Progressive}    
                              & CTC &  -  & 25.73 & -   \\
                              & \;+ XCTC &  14.85  & 26.60  & 27.04  \\
                              & \;\;\;+ InterCTC & 18.90 & 26.49 & 27.57 \\
                              & \;\;\;\;\;+ PAE & 18.32 & 26.60 & 27.37 \\
                              & \;\;\;\;\;\;\;+ CLM & 20.87 & 27.19 & \textbf{27.80} \\
                              & \;\;\;\;\;\;\;\;\;+ KD Data & \textbf{25.17} & \textbf{27.47} & 27.64 \\
                              
    \specialrule{0em}{1pt}{1pt}
    \cdashline{1-5}
    \specialrule{0em}{1pt}{2pt}
    
    \multirow{6}{*}{Synchronous} 
                              & CTC           & -   &  26.73 & -   \\
                              & \;+ XCTC        &  14.24  & 26.60  &  27.24 \\
                              & \;\;\;+ InterCTC & 15.10 & 26.83 & 27.57 \\
                              & \;\;\;\;\;+ PAE & 15.35 & 27.17 & 27.81 \\
                              & \;\;\;\;\;\;\;+ CLM & 16.93 & 27.75 & \textbf{28.35} \\
                              & \;\;\;\;\;\;\;\;\;+ KD Data & \textbf{22.21} & \textbf{27.88} & 27.89 \\

    \bottomrule
  \end{tabular}
  \caption{BLEU scores on MuST-C En-De tst-COMMON test set under different inference methods.}
  \label{mustc_ende}
\end{table}
\begin{table*}[t!]
  \centering
  \resizebox{\textwidth}{!}{
  \tiny
  \begin{tabular}{lcccccccccccc}
    \toprule
    \specialrule{0em}{2pt}{2pt}
    \multirow{2}*{Model} & \multicolumn{2}{c}{External Data} & \multirow{2}*{De} & \multirow{2}*{Es} & \multirow{2}*{Fr} & \multirow{2}*{It} & \multirow{2}*{Nl} & \multirow{2}*{Pt} & \multirow{2}*{Ro} & \multirow{2}*{Ru} & \multirow{2}*{Ja} & \multirow{2}*{Avg.} \\
    \cmidrule(lr){2-3} 
    & Unlabeled & labeled & & & & & &  \\

    \Xhline{0.6pt}
    \specialrule{0em}{0.6pt}{0.8pt}
    
    CMOT$^{\prime}$23 \cite{zhou2023cmot} &\usym{2713} & \usym{2613} & 27.0 & 31.1 & 37.3 & 26.9 & 31.2 & 32.7 & 25.3 & 17.9 & - & 28.7 \\
    CRESS$^{\prime}$23 \cite{DBLP:conf/acl/FangF23a} & \usym{2713} & \usym{2613} & 27.2 & 31.9 & 37.8 & 27.3 & 31.6 & 33.0 & 25.9 & 18.7 & - & 29.2 \\
    
    CTC Aug$^{\prime}$23 \cite{nast} & \usym{2613} & \usym{2613} & 26.9 & 31.5 & 38.1 & 27.4 & 31.9 & 33.4 & 25.8 & 18.7 & 16.1 & 29.2 \\
    \quad + KD Data & \usym{2613} & \usym{2613} & 27.7 & 31.6 & 39.5 & 27.5 & 32.3 & 33.7 & 26.6 & 18.7 & 16.4 & 29.7 \\
    
    Siamese-PT$^{\prime}$23 \cite{pmlr-v202-le23a} & \usym{2613} & \usym{2613} & 27.9 & 31.8 & 39.2 & 27.7 & 31.7 & 34.2 &  \textbf{27.0} & 18.5 & - & 29.8 \\

    \specialrule{0em}{0.4pt}{0.4pt}
    \cdashline{1-13}[2pt/3pt]
    \specialrule{0em}{0.5pt}{1pt}
    
    XSTNet$^{\prime}$21 \cite{ye2021end} & \usym{2713} & \usym{2713} & 27.8 & 30.8 & 38.0 & 26.4 & 31.2 & 32.4 & 25.7 & 18.5 & - & 28.8 \\
    
    STEMM$^{\prime}$22 \cite{fang2022stemm} & \usym{2713} & \usym{2713} & 28.7 & 31.0 & 37.4 & 25.8 & 30.5 & 31.7 & 24.5 & 17.8 & - & 28.4 \\

    ConST$^{\prime}$22 \cite{ye2022cross} & \usym{2713} & \usym{2713} & 28.3 & 32.0 & 38.3 & 27.2 & 31.7 & 33.1 & 25.6 & 18.9 & - & 29.4 \\

    FCCL$^{\prime}$23 \cite{zhang2023improving} & \usym{2713} & \usym{2713} & 29.0 & 31.9 & 38.3 & 27.3 & 31.6 & 32.7 & 26.8 & \textbf{19.7} & - & 29.6 \\
        
    M$^3$ST$^{\prime}$23 \cite{cheng2023mst} & \usym{2713} & \usym{2713} & 29.3 & 32.4 & 38.5 & 27.5 & 32.5 & 33.4 & 25.9 & 19.3 & - & 29.9 \\
    
    CMOT$^{\prime}$23 \cite{zhou2023cmot} & \usym{2713} & \usym{2713} & 29.0 & 32.8 & 39.5 & 27.5 & 32.1 & 33.5 & 26.0 & 19.2 & - & 30.0 \\
        
    CRESS$^{\prime}$23 \cite{DBLP:conf/acl/FangF23a} & \usym{2713} & \usym{2713} & \textbf{29.4} & \textbf{33.2} & \textbf{40.1} & 27.6 & 32.3 & 33.6 & 26.4 & \textbf{19.7} & - & \textbf{30.3} \\

    \specialrule{0em}{0.4pt}{0.4pt}
    \cdashline{1-13}[2pt/3pt]
    \specialrule{0em}{0.5pt}{1pt}
    
    BiL-CTC+ (Ours) & \usym{2613} & \usym{2613} & 27.8 & 32.4 & 38.9 & 28.1 & 32.7 & 33.9 & 26.7 & 19.2 &  16.8   & 30.0\\
    
    \quad + Re-scoring & \usym{2613} & \usym{2613} &   28.4    &    32.0   &   39.5    &  \textbf{28.2}     &  \textbf{33.0}     &   \textbf{34.3}    &  \textbf{27.0}    &   19.6   &  \textbf{17.0}    & \textbf{30.3}     \\
    
    \bottomrule
  \end{tabular}
}
  \caption{BLEU scores on MuST-C corpora.}
  \label{mustc_all}
\end{table*}

We investigate the effects of both progressive and synchronous strategies on the MuST-C En-De ST benchmark.
Our evaluations also extend to leveraging sequence-level knowledge distillation \cite{kim2016sequence} data to facilitate the cross-lingual learning of XCTC \cite{nast}.
As shown in Table \ref{mustc_ende}, the progressive approach consistently outperforms its synchronous counterpart in terms of CTC prediction.
We speculate that this could be due to linguistic interference introduced by synchronous prediction, resulting in code-switching outputs.

Conversely, synchronous prediction demonstrates remarkable superiority when utilizing Attn-only inference, validating our hypothesis that it encourages language-agnostic representations and thereby enhances semantic understanding. 
The employment of the re-scoring method to combine CTC and attention-based predictions further improves the performances.

Intriguingly, our experiments reveal that KD data leads to significant performance improvements for both CTC-only and Attn-only decoding schemes. 
However, these gains are noticeably marginal when applied to re-scoring inference.
This finding suggests that a more diverse prediction distribution in CTC and attention could substantially benefit the re-scoring process. 
Consequently, in subsequent experiments, we choose to train the model employing the synchronous BiL-CTC approach on the original dataset.
\begin{table}[t!]
  \centering
  \footnotesize
  \begin{tabular}{lrrrr}
    \toprule
    \multirow{2}*{Model} & \multicolumn{2}{c}{dev} & \multicolumn{2}{c}{test} \\
    \cmidrule(lr){2-3}\cmidrule(lr){4-5}
    \specialrule{0em}{1pt}{1pt}
    & clean & other & clean & other \\
    \midrule
    CTC & 7.54 & 18.08 & 7.83 & 17.97 \\
    \;+ InterCTC & 7.26 & 17.94 & 7.64 & 18.20 \\ 
    \;\;\;+ XCTC & 7.19 & 17.37 & 7.40 & 17.65 \\
    \;\;\;\;\;+ PAE & 7.29 & 17.76 & 7.37 & 17.48  \\
    \;\;\;\;\;\;\;+ CLM & 6.80 & 17.18 & 7.20 & \textbf{17.03} \\
    \;\;\;\;\;\;\;\;\;+ Re-scoring & \textbf{6.60} & \textbf{16.90} & \textbf{6.63} & 17.13 \\
    
    \bottomrule
  \end{tabular}
  \caption{WER on LibriSpeech 100h dataset. The Attn-only inference is employed by default.}
  \label{asr}
\end{table}
\subsection{Main Results on ST}

Table \ref{mustc_all} presents the comprehensive results across the whole MuST-C corpora. 
Notably, our proposed BiL-CTC+ approach eliminates the necessity for complex design decisions and establishes a new state-of-the-art performance in scenarios without external data.
Our approach even surpasses a majority of existing methods that are trained on expansive external datasets.

A potential concern may arise regarding the effects of synchronous prediction when dealing with linguistically distant language pairs. 
To validate this, we extend our method to the MuST-C v2 En-Ja dataset, thereby assessing the capability to handle disparate linguistic attributes between transcript and translation.
Impressively, the BiL-CTC+ approach yields remarkable performance gains compared with the strong benchmarks in the literature. 
These findings underscore both the robustness and the broad applicability of our proposed methodology.

\subsection{Main Results on ASR}

\begin{table}[t!]
  \centering
  \footnotesize
  \begin{tabular}{llrrrr}
    \toprule
    \multirow{2}*{Model} & \multirow{2}*{Inference} & \multicolumn{2}{c}{dev} & \multicolumn{2}{c}{test} \\
    \cmidrule(lr){3-4}\cmidrule(lr){5-6}
    \specialrule{0em}{1pt}{1pt}
     & & clean & other & clean & other \\
    \midrule
    \multirow{3}{*}{Baseline}
    & CTC & 2.61 & 6.74 & 2.84 & 6.78 \\ 
    & Attn & 3.42 & 6.71 & 3.65 & 6.90 \\
    & Re-scoring & 3.31 & 6.59 & 3.46 & 6.68 \\
    \specialrule{0em}{0.4pt}{0.4pt}
    \cdashline{1-6}
    \specialrule{0em}{0.5pt}{1pt}
    \multirow{3}{*}{BiL-CTC+}   
    & CTC & \textbf{2.37} & 6.16 & \textbf{2.55} & 6.09  \\ 
    & Attn & 2.74 & 5.95 & 2.87 & 6.04 \\
    & Re-scoring & 2.47 & \textbf{5.86} & 2.61 & \textbf{5.94} \\
    \bottomrule
  \end{tabular}
  \caption{WER on LibriSpeech 960h dataset.}
  \label{960}
\end{table}

Contrary to the natural matching of BiL-CTC with the ST task, cross-lingual modeling may be irrelevant or even harmful to the performance of ASR. 
To empirically evaluate this, we conduct experiments on the LibriSpeech 100h and 960h datasets, as detailed in Tables \ref{asr} and \ref{960}. 
Intriguingly, our BiL-CTC+ method demonstrates consistent performance gains, which contradicts the intuition. 
We conjecture that the introduction of bilingual prediction into the encoder also enhances semantic accuracy in ASR, thus preventing the model from excessively focusing on pronunciation aspects.
We will investigate this counterintuitive finding in future work.

\section{CONCLUSIONS}

In this paper, we introduce BiL-CTC, which leverages dual CTC objectives to learn bilingual prediction. 
We propose an innovative synchronous strategy that guides the model to generate language-agnostic representations, thereby enhancing its capability for semantic understanding. 
Augmented with recent advances in CTC applications, our BiL-CTC+ model establishes new state-of-the-art performance on the MuST-C dataset. 
Intriguingly, our approach also yields significant improvements in the ASR task, thereby highlighting the broad applicability of our method.

Our findings expose the previously untapped, remarkable potential of bilingual learning paradigms.
Owing to its simplicity, effectiveness, and versatile applicability, the BiL-CTC+ framework has the potential to be a compelling new paradigm in both ASR and ST.

\clearpage

\bibliographystyle{IEEEbib}
\bibliography{strings,refs}

\end{document}